\documentclass[lettersize,journal]{IEEEtran}
\usepackage{amsmath,amsfonts}
\usepackage{algorithmic}
\usepackage{algorithm}
\usepackage{array}
\usepackage{amssymb}

\usepackage{booktabs}
\usepackage{bbding}
\usepackage{bm}

\usepackage{cite}

\usepackage[caption=false,font=normalsize,labelfont=sf,textfont=sf]{subfig}
\usepackage{textcomp}
\usepackage{stfloats}
\usepackage{url}
\usepackage{verbatim}

\usepackage{graphicx}

\usepackage{multirow}

\usepackage{float}

\usepackage{microtype}
\usepackage{wasysym}

\usepackage{pifont}

\usepackage{ulem}
\usepackage{authblk}

\DeclareRobustCommand*{\IEEEauthorrefmark}[1]{%
    \raisebox{0pt}[0pt][0pt]{\textsuperscript{\footnotesize\ensuremath{#1}}}}

\begin{document}

\title{Multi-scale Information Sharing and Selection Network with Boundary Attention for Polyp Segmentation}

\author{
\IEEEauthorblockN{
Xiaolu Kang,
Zhuoqi Ma,
Kang Liu,
Yunan Li,
Qiguang Miao,
IEEE Senior Member} \\
\IEEEauthorblockA{\IEEEauthorrefmark{a}School of Computer Science and Technology, Xidian University}}

\renewcommand*{\Affilfont}{\small\it} 

\markboth{Journal of \LaTeX\ Class Files,~Vol.~XX, No.~XX, January~2024}%
{Shell \MakeLowercase{\textit{et al.}}: A Sample Article Using IEEEtran.cls for IEEE Journals}


\maketitle

\begin{abstract}


Polyp segmentation for colonoscopy images is of vital importance in clinical practice. It can provide valuable information for colorectal cancer diagnosis and surgery. While existing methods have achieved relatively good performance, polyp segmentation still faces the following challenges: (1) Varying lighting conditions in colonoscopy and differences in polyp locations, sizes, and morphologies. (2) The indistinct boundary between polyps and surrounding tissue. To address these challenges, we propose a Multi-scale information sharing and selection network (MISNet) for polyp segmentation task. We design a Selectively Shared Fusion Module (SSFM) to enforce information sharing and active selection between low-level and high-level features, thereby enhancing model's ability to capture comprehensive information. We then design a Parallel Attention Module (PAM) to enhance model's attention to boundaries, and a Balancing Weight Module (BWM) to facilitate the continuous refinement of boundary segmentation in the bottom-up process. Experiments on five polyp segmentation datasets demonstrate that MISNet successfully improved the accuracy and clarity of segmentation result, outperforming state-of-the-art methods.
\end{abstract}


\section{Introduction}

\IEEEPARstart{C}{olorectal} cancer (CRC) stands as one of the most prevalent cancers in the world. The disease is mainly attributed to the malignant growth of bulky tissue known as polyps in the colon or rectum. The standard approach for CRC diagnosis involves colonoscopy examinations, allowing for the visualization of the location and appearance of polyps. Early detection and treatment of rectal polyps can effectively prevent the occurrence of rectal cancer. However, with increasingly growing medical pressure, manual diagnosis is time-consuming, labor-intensive, and less stable. Hence, accurate and efficient polyps segmentation is of crucial significance in clinical practice.

Many research have been conducted in polyp segmentation, demonstrating notable progress. Some methods introduce fully convolutional neural networks(FCN) \cite{brandao2017fully, akbari_polyp_2018, wichakam_real-time_2018} for pixel-level prediction of polyps in colonoscopy. Although FCN-based methods are highly efficient, the direct up-sampling operation could lead to blurred results due to the loss of details. Addressing this problem, UNet \cite{ronneberger_u-net_2015} proposed a U-shaped encoder-decoder structure for medical image segmentation. UNet fuses different level of features through skip connections, enabling more reasonable restoration of details. Its variants \cite{zhou2018unet++, jha_resunet_2019} also exhibited remarkable performance for polyp segmentation. Still, these methods overlooked the valuable boundary information. 

Some research have explored solutions to this issue via introducing boundary information into polyp segmentation. Psi-Net \cite{murugesan2019psi} employs a parallel decoder for jointly training in three tasks: mask, contour, and distance map. Meanwhile, a new joint loss function is proposed, achieving better results by retaining more boundary information. However, the integration of too many tasks has led to shortcomings in this model when it comes to extracting the boundary relationships between polyps and surrounding tissues. SFANet \cite{fang_selective_2019} proposes a area and boundary constraints with additional edge supervision. However, the limited expressive capability of the model encoder for polyp image features results in poor internal coherence in image segmentation and unclear segmentation edges.
PraNet \cite{fan_pranet_2020} utilizes reverse attention \cite{ferrari_reverse_2018} to acquire additional boundary information. However, reverse attention tends to focus more on the background area, introducing noticeable noise in the predicted results.

Unfortunately, in aforementioned methods, low-level features are often overlooked as they are considered to contribute less to the network compared to high-level features but with high computation cost. However, these features which capturing details like edges and textures, can offer valuable local fine-grained information for the identification of boundaries and subtle structures, thereby complementing and assisting in segmentation tasks. Besides, the exploitation of boundary information still needs refinement to enhance segmentation precision and clarity. Therefore, we aim to tackle two challenges in polyp segmentation: (1) Exploring effective and efficient ways to incorporate low-level features for mining boundary cues. (2) Refining boundary information exploitation strategy.


In this paper, we propose a novel neural network, called Multi-scale Information Sharing and Selection Network (MISNet) for the polyp segmentation task. First, MISNet adopt a Selectively Shared Fusion Module (SSFM) to improve model's ability to capture multi-scale contextual information, thus the generated initial guidance map can address the scale variation challenges of polyps. Second, a set of Parallel Attention Modules (PAM) are introduced to further mine the polyp boundary information. Third, we use Balancing Weight Module (BWM) to adaptively incorporate the low-level feature, boundary attention and guidance map, enabling our model to continuously refine boundary details in the bottom-up flow of the network. With BWM, the low-level feature would serve as explicit guide to further improve the boundary segmentation. Benefiting from these well-designed modules, the proposed network demonstrates enhanced accuracy and clarity for polyp segmentation.

In conclusion, our main contributions are summarized as follows:

(1) We propose a Selectively Shared Fusion Module (SSFM) to enforce information sharing and active selection between features at different scales, enabling the model to capture boundary details as well as global context information. 

(2) We present a new method with Parallel Attention Module (PAM) and Balancing Weight Module (BWM) to effectively extract and exploit boundary information for enhancing polyp segmentation accuracy and clarity.

(3) Extensive experiments on five polyp segmentation datasets demonstrate that the proposed network outperforms state-of-the-art methods. Meanwhile, a comprehensive ablation studies validate the effectiveness of key components in our proposed model.

\section{Related Work}

\subsection{Medical Image Segmentation}

Recently, deep learning has demonstrated remarkable performance for precise medical image segmentation. UNet \cite{ronneberger_u-net_2015} achieves the segmentation process through a symmetric encoder-decoder structure. Following this approach, various types of improvement structures have been primarily designed for segmentation research.

The first improvement focuses on skip connections. UNet++ \cite{zhou2018unet++} enhances the fusion of multi-scale features by using a series of nested dense skip connections on both the encoder and the decoder. U2-Net \cite{qin2020u2} defines a nested U-shaped structure and introduces Residual U-blocks, achieving the capture of context information at different scales.
The second enhancement involves utilizing different types of backbone. ResUNet \cite{jha_resunet_2019} introduces ResNet \cite{7780459} and enhances learning performance by fitting residuals by adding skip connections. ResUNet++ \cite{jha_resunet_2019} improves ResUNet by integrating efficient components into the UNet structure. R2U-Net \cite{alom2018recurrent} combines the advantages of UNet, residual networks, and Recurrent CNNs(RCNNs) to design a recurrent residual convolutional neural network, which can perform the rich feature representation and segment the target successfully.
Next is the incorporation of various mechanisms into UNet, such as the attention mechanism in Attention UNet \cite{oktay_attention_2018}. It incorporates attention gates (AG) into skip connections, allowing the network to emphasize the salient features of the specific target for effective segmentation. Attention UNet++ \cite{li_attention_2020} further enhances UNet++ by adding attention gates (AG) between nested convolution blocks, allowing features at different levels to selectively focus on their respective tasks. ERDUnet \cite{li2023erdunet} addresses the challenge of extracting global contextual features by enhancing UNet, aiming to improve segmentation accuracy while saving parameters.

After that, transformer-based \cite{vaswani2017attention} methods have introduced new ideas for medical image segmentation. Medical Transformer (MedT) \cite{valanarasu2021medical} based on Transformer introduces a gated axial self-attention mechanism and local global training strategy (LoGo) to learn image features automatically. SwinUNet \cite{cao2022swin} employs a hierarchical Swin Transformer \cite{liu2021swin} for feature extraction and designs a decoder based on the symmetric Swin Transformer with patch expansion layers to perform upsampling operations. TransUNet \cite{chen_transunet_2021} combines the strengths of UNet and Transformer, exhibiting robust performance in medical image segmentation. MSCAF-Net \cite{liu2023mscaf} adopts the improved Pyramid Vision Transformer (PVTv2) model as its backbone, refining features at each scale and achieving a comprehensive interaction of multi-scale information for accurate segmentation.

\subsection{Polyp Segmentation}

Deep learning has been widely applied in polyp segmentation. Brandao  et al. \cite{brandao2017fully} are pioneers in utilizing a Fully Convolutional Neural Network \cite{shelhamer_fully_2017} (FCN) to segment polyps in colonoscopy. Wicakam et al. \cite{wichakam_real-time_2018} propose a fully compressed convolutional network by improving the FCN-8s network, enabling real-time polyp segmentation and improving segmentation performance. Wickstrm et al. \cite{wickstrom2020uncertainty} propose an advanced architecture that combines FCN-8s and SegNet \cite{badrinarayanan2017segnet}, introducing batch normalization and dropout to improve model generalization and estimate model uncertainty. 

However, the up-sampling process of FCN-based methods often results in the loss of detailed information.
To solve this problem, UNet++ and ResUNet++ based on UNet serve for polyp segmentation successfully. PolypMixer \cite{shi2022polyp} is a model based on MLP that flexibly handles input scales of polyps and models long-term dependencies for precise and efficient polyp segmentation. However, these approaches often pay insufficient attention to valuable boundary details.

Several approaches have actively explored solutions to this problem. SFANet \cite{fang_selective_2019} introduces a boundary-sensitive loss and employs a shared encoder with two mutually constrained decoders to select and aggregate polyp features at different scales. PsiNet \cite{murugesan2019psi} employs parallel decoders for joint training of three tasks and proposes a new joint loss function that achieves improved results after retaining more boundary information. However, it still exhibits limitations in handling boundaries due to the integration of multiple tasks. In addition, PraNet \cite{fan_pranet_2020} utilizes reverse attention \cite{ferrari_reverse_2018} to acquire additional boundary cues. CaraNet \cite{lou_caranet_2023} proposes a contextual axial reverse attention network by adding axial attention to the reverse attention module and using the channel feature pyramid (CFP) module to improve the segmentation performance of small targets. ACSNet \cite{zhang2020adaptive} leverages local and global contextual features to achieve layer-wise feature complementarity and refine predictions for uncertain regions. However, it extracts limited information from the last feature generated by the encoder for guidance and focusing attention entirely on boundaries might make it challenging for the model to segment the polyp region. CCBANet \cite{nguyen_ccbanet_2021} introduces attention to the foreground, background, and boundary regions to ensure that the model covers the segmentation area with attention as much as possible. MSNet \cite{zhao_automatic_2021} introduces a multi-scale subtraction network and comprehensively supervises features to capture more details and structural cues for accurate polyp localization and edge refinement. BDG-Net \cite{qiu2022bdg} proposes a boundary-distribution-guided segmentation network to aggregate high-level features and generate a boundary distribution map, which successfully segments polyps at different scales. DCNet \cite{yue2023dual} explores candidate objects and additional object-related boundaries by constraining object regions and boundaries. It segments polyps in a coarse-to-fine manner.

\section{Methodology}

\subsection{Overview}

\begin{figure*}  
\centering 
\includegraphics[width = \textwidth]{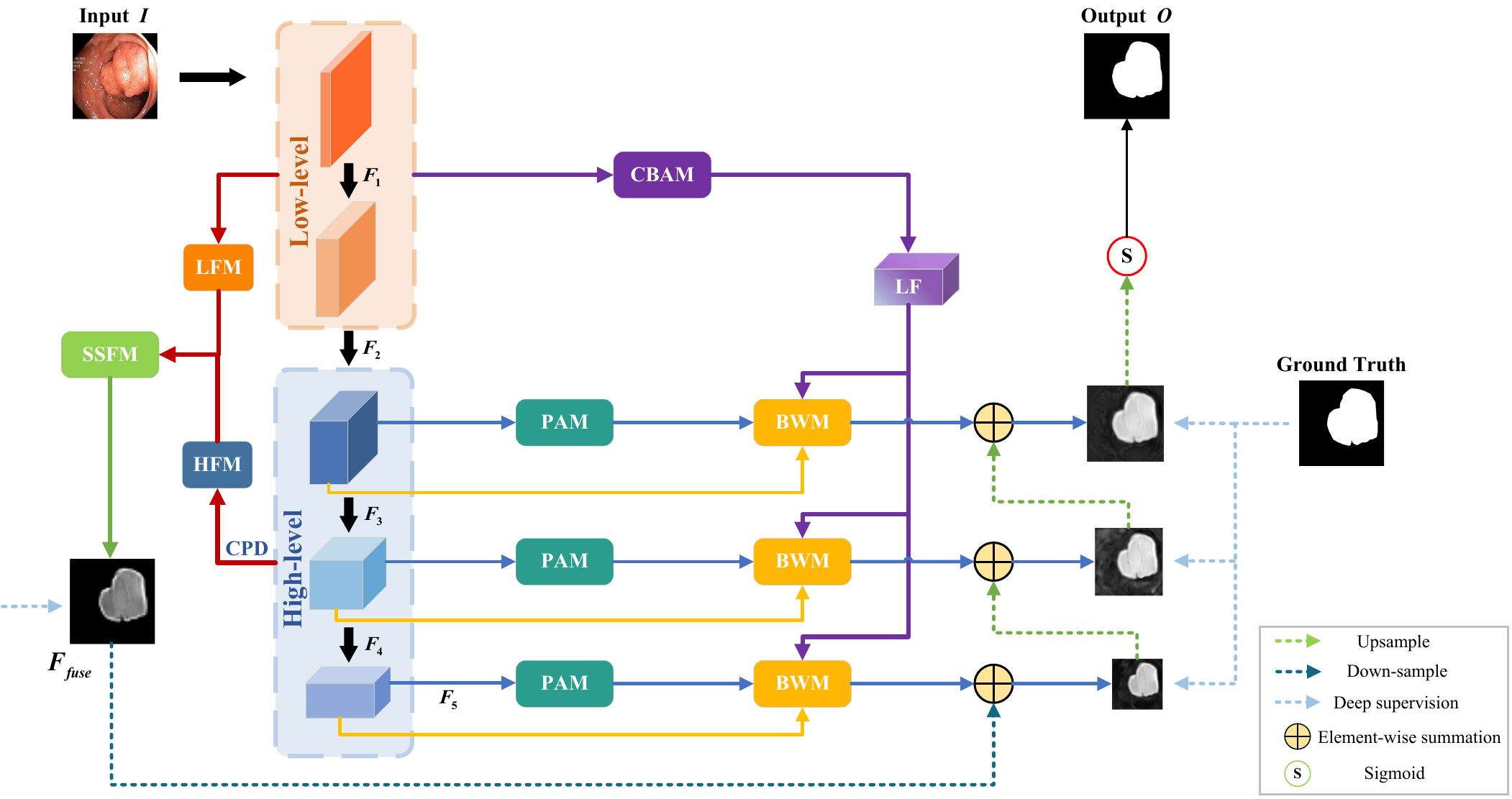} 
\caption{Overview of the proposed method.}
\label{fig_1}
\end{figure*}

The overall architecture of MISNet is illustrated in Figure \ref{fig_1}. For the input colonoscopy image $I$, we first use Res2Net \cite{gao2019res2net} to extract features at five different scales. To improve computation efficiency, we use Low-level Fusion Module (LFM) to integrate shallow features extracted from the first two blocks of backbone, and use High-level Fusion Module (HFM) to integrate deep features extracted from the last three blocks of backbone. 

MISNet first use Selectively Shared Fusion Module (SSFM) to adaptively aggregate low-level and high-level features and generate an initial map $F_{fuse}$ for the following process. Then, we leverage Parallel Attention Modules (PAM) to emphasize local boundary information. Subsequently, the features of the current layer, the boundary attention from PAM, and the explicit boundary cues mined from low-level features are adaptively combined with Balancing Weight Module (BWM) to refine polyp segmentation in the bottom-up flow of the network. We elaborate each component in below.

\subsection{Low-level Feature Fusion and High-level Feature Fusion}

\begin{figure*}  
\centering 
\includegraphics[width = \textwidth]{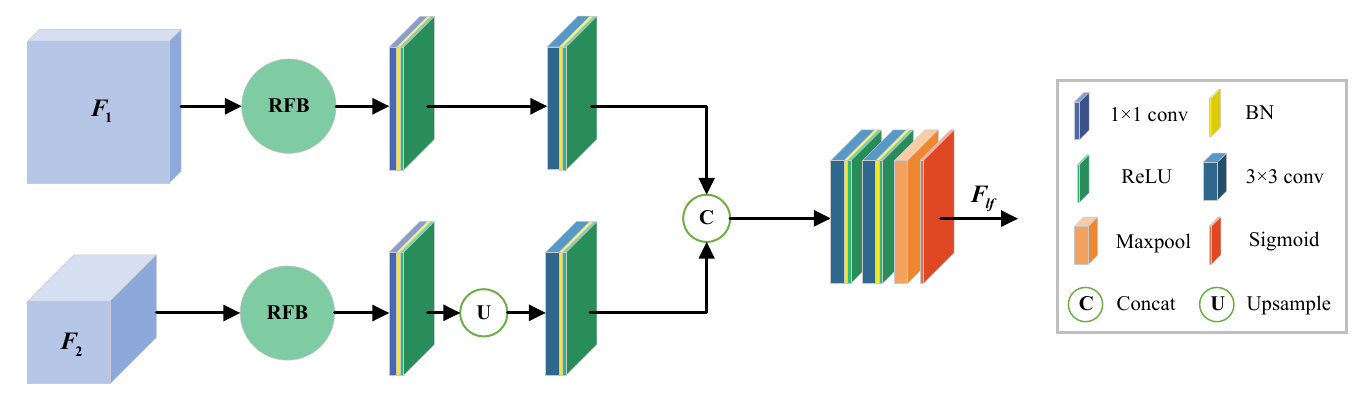} 
\caption{Illustration of Low-level Fusion Module (LFM).}
\label{fig_2}
\end{figure*}

With the significance of preserving valuable local fine-grained information for polyp boundary identification, we introduce a Low-level Fusion Module (LFM). This module facilitates the exploitation of low-level features, enhancing the model's ability to extract precise boundary information.


The low-level fusion module is shown in Figure \ref{fig_2}. Specifically, shallow features ${F_1}$ and ${F_2}$ generated by the first two blocks of the backbone are firstly processed by Receptive Field Block (RFB) \cite{liu2018receptive} to expand the receptive field. Then, the channel number of ${F_1}$ and ${F_2}$ is adjusted by $1 \times 1$ convolutional layers respectively, after which ${F_2}$ will be upsampled to the same resolution as that of ${F_1}$. Then, these two features are passed through $3 \times 3$ convolutional layers then concatenated to generate a aggregated feature. The aggregated feature is then processed with two $3 \times 3$ convolution layers and downsampled to match with the the selective shared fusion module.
 
The fused low-level features ${F_{lf}}$ serve two purposes within the network. Firstly, they contribute as inputs to SSFM, aiding in the sharing and adaptive selection of shallow features, thereby enhancing the accuracy of the initial guidance map. Secondly, these features explicitly guide the boundary refinement in the decoding process, facilitating clarity enhancement in the generation of segmentation results.


For the fusion of deep features, we adopt the Cascaded Partial Decoder \cite{wu2019cascaded}. Specifically, the high-level features $\{ {F_i},i = 3,4,5\}$ produced by the last three blocks of the backbone are first expanded in receptive field through RFB. Then, the high-level features are aggregated using the Cascaded Partial Decoder $cpd\left(\cdot\right)$, ${F_{hf}} = cpd({F_3},{F_4},{F_5})$.




 \begin{figure*}  
\centering 
\includegraphics[width = \textwidth]{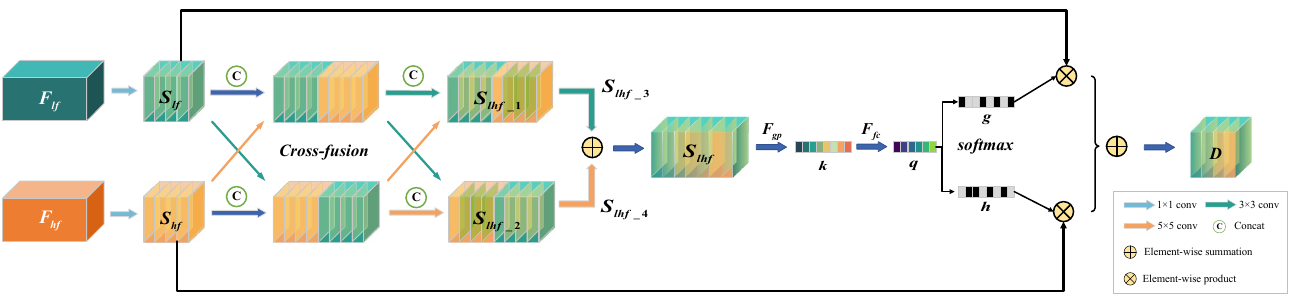} 
\caption{Illustration of Selectively Shared Fusion Module (SSFM).}
\label{fig_3}
\end{figure*}

\subsection{Selectively Shared Fusion Module}

As demonstrated by most image recognition research, high-level features capture abstract and high-level semantic information, providing a more global context, whereas low-level features contribute to capturing details and local information. For segmentation tasks, high-level features enable the model to better understand the complex structures and relationships within the image, enhancing the model's ability to recognize and locate objects, while low-level features provide finer details, assisting the model in more accurately locating and segmenting the boundaries of objects.

Thus, appropriate integration of features at different scales is crucial for providing comprehensive information in image segmentation tasks. In conventional "U-shaped" network structures, features of the same scale are typically fused through skip connections in a single stream, which is not conducive for information interaction. Motivated by these observations, we propose a novel Selectively Shared Fusion Module (SSFM) to enforce information sharing and active selection between features at different levels. 




The structure of SSFM is illustrated in Figure \ref{fig_3}. Firstly, the channels of fused low-level feature ${F_{lf}}$ and high-level feature ${F_{hf}}$ are reduced with $1 \times 1$ convolution to obtain the squeezed feature $S_{lf} \in \mathbb{R}^{W \times H \times C}$ and $S_{hf} \in \mathbb{R}^{W \times H \times C}$. Then, we employ cross-fusion to share information between $S_{lf}$ and $S_{hf}$. Specifically, $S_{lf}$ and $S_{hf}$ are first cross-concatenated, and then passed through two separate convolutional layers with different kernels to obtain more comprehensive feature representation:
\begin{equation}
\left\{ \begin{array}{l}
{S_{lhf\_1}} = {C_{3 \times 3}}(concat({S_{lf}},{S_{hf}}))\\
{S_{lhf\_2}} = {C_{5 \times 5}}(concat({S_{hf}},{S_{lf}})),
\end{array} \right.
\end{equation}where ${C_{3\times3}}$ and ${C_{5\times5}}$ denote a $3\times3$ convolutional layer and a $5\times5$ convolutional layer followed by a BN layer and a ReLU activation function. 
Then we obtain ${S_{lhf\_3}}$ and ${S_{lhf\_4}}$ by repeating cross-fusion:

\begin{equation}
\left\{ \begin{array}{l}
{S_{lhf\_3}} = {C_{3 \times 3}}(concat({S_{lhf\_1}},{S_{lhf\_2}}))\\
{S_{lhf\_4}} = {C_{5 \times 5}}(concat({S_{lhf\_2}},{S_{lhf\_1}})),
\end{array} \right.
\end{equation}With two rounds of cross-fusion, the low-level and high-level features complete the circulation of information. 

Subsequently, we allow the model to adaptively select the essential information from low-level features and high-level features. We first apply an element-wise summation operation to the features from these two branches: 
\begin{equation}
    {S_{lhf}} = {S_{lhf\_3}} + {S_{lhf\_4}},
\end{equation}Global average pooling is then employed to embed the global information.
In the fused feature ${S_{lhf}}$, the channel-wise mean value $k$ is computed, $k \in \mathbb{R}^C$. For each channel, we have:

\begin{equation}
     {k_c} = {F_{gp}}({S_{lhf\_c}}) = \frac{1}{{H \times W}}\sum\limits_{i = 1}^H {\sum\limits_{j = 1}^W {{S_{lhf\_c}}(i,j)} } ,
\end{equation}
Then, we use a $C \times d$ fully connected layer $FC_{C \times d}$ to obtain a relatively dense feature $q \in \mathbb{R}^{d \times 1}$ by reducing the dimension of $k$:
\begin{equation}
    q = \delta (\beta (FC_{C \times d}(k)),
\end{equation}where $\delta $ denotes the sigmoid function and $\beta $ denotes the BN.

 
\begin{equation}
    d = \max (C/r,L),
\end{equation}
where $r$ denotes the reduction ratio, $L$ denotes the minimal value of $d$ ($L$ is 16 in our experiments). 


Following that, we compute the attention for low-level feature and high-level feature:
\begin{equation}
g_c=\frac{e^{G_c q}}{e^{G_c q}+e^{H_c q}}, \quad h_c=\frac{e^{H_c q}}{e^{G_c q}+e^{H_c q}},
\end{equation}
where $G, H \in \mathbb{R}^{C \times d}$ are learnable weights, and $g$, $h$ denote the soft attention for ${S_{lf}}$ and ${S_{hf}}$. $G_c \in \mathbb{R}^{1 \times d}$, $H_c \in \mathbb{R}^{1 \times d}$ is the c-th row of $G$ and $H$. $g_c$, $h_c$ is the c-th element of $g$ and $h$.

Finally, the information from low-level feature and high-level feature are adaptively selected by multiplying ${S_{lf}}$ and ${S_{hf}}$ with attention weight and obtain the initial guidance map $D$: 
\begin{equation}
D_c=g_c \cdot S_{lf_{-} c}+h_c \cdot S_{hf_{-} c}, \quad g_c+h_c=1,
\end{equation}
where $D = [{D_1},{D_2},...,{D_C}]$ , $D_c \in \mathbb{R}^{H \times W}$.

With information sharing and adaptive selection between low-level and high-level feature, SSFM can attend to both local details and global contextual information, thus generate more accurate initial guidance map. 



\subsection{Parallel Attention Module}

\begin{figure}  
\centering 
\includegraphics[width = 0.48\textwidth]{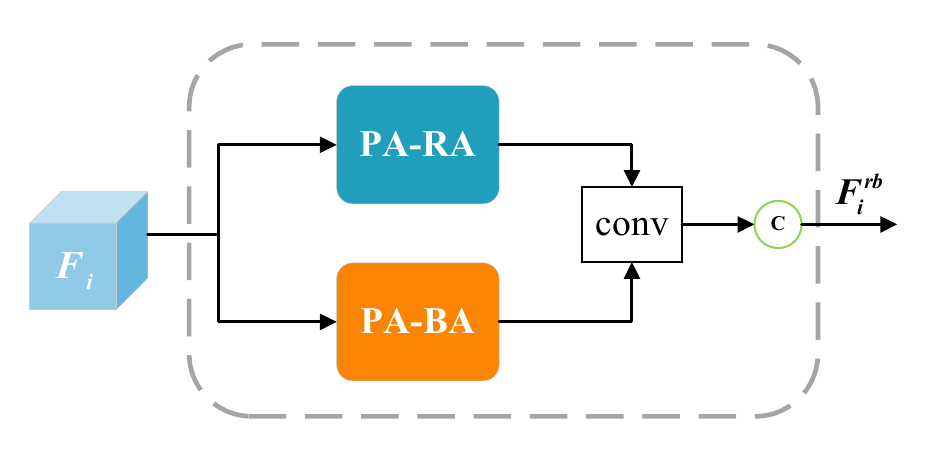} 
\caption{Illustration of Parallel Attention Module (PAM).}
\label{fig_4}
\end{figure}

The initial guidance map generated through SSFM can provide a coarse area and location of the polyp. Still, it lacks accurate identification and localization of object boundaries. To address this issue, we introduce a parallel attention module to progressively refine the boundary cues of three parallel high-level features during the decoding stage. 



\begin{figure}  
\centering 
\includegraphics[width = 0.48\textwidth]{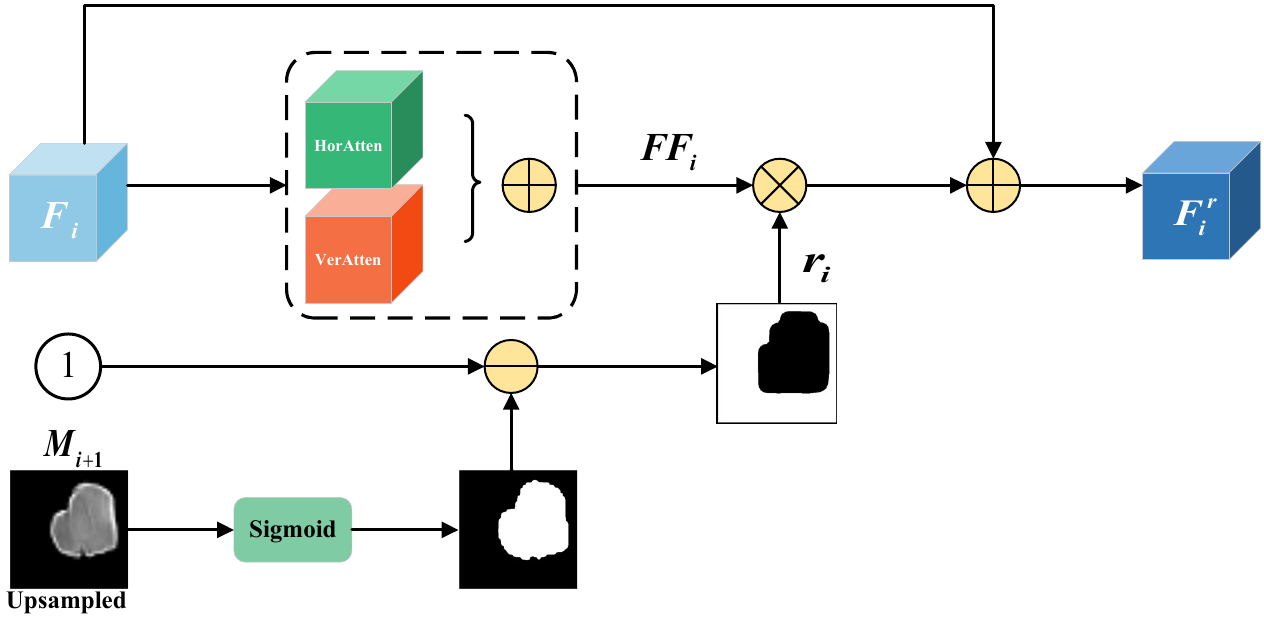} 
\caption{Illustration of Parallel Axial Reverse Attention (PA-RA).}
\label{fig_5_a}
\end{figure}

\begin{figure}  
\centering 
\includegraphics[width = 0.48\textwidth]{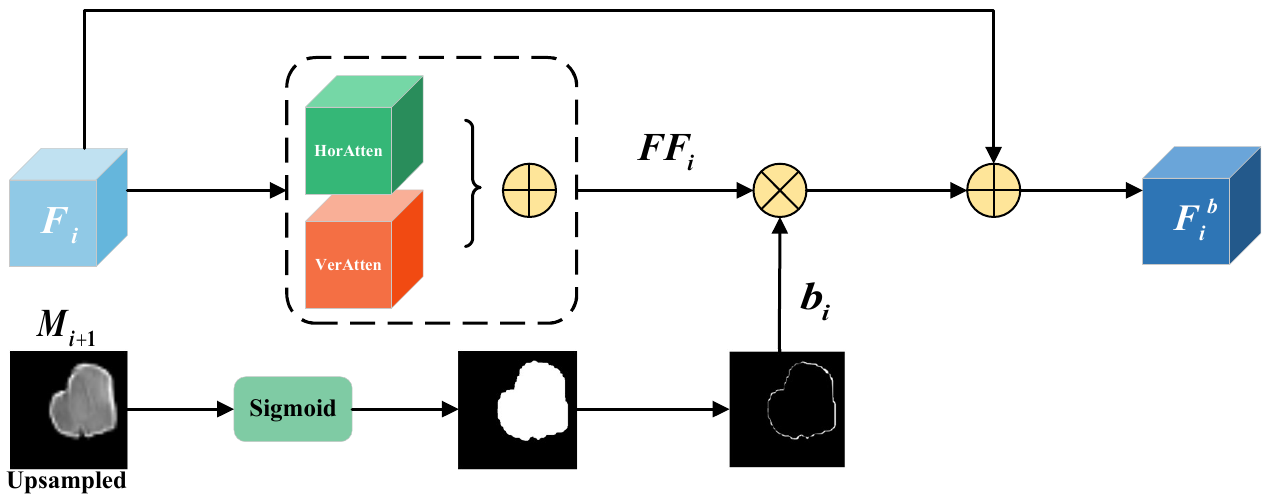} 
\caption{Illustration of Parallel Axial Boundary Attention (PA-BA).}
\label{fig_5_b}
\end{figure}

The structure of Parallel Attention Module is illustrated in Figure \ref{fig_4}. PAM consist of two parallel attention branches: Parallel Axial Reverse Attention (PA-RA) and Parallel Axial Boundary Attention (PA-BA). PA-RA learns the relationship between background and target areas, and PA-BA further address the attention to boundary information. The structure of PA-RA and PA-BA is demonstrated in Figure \ref{fig_5_a} and \ref{fig_5_b}.







\subsubsection{PA-RA}




As our backbone is pre-trained on ImageNet, high-level features may prioritize polyp regions with higher response values, potentially ignoring boundary details. Therefore, we introduce Parallel Axial Reverse Attention to capture the object boundary features more accurately. The erasure strategy within Parallel Axial Reverse Attention refines imprecise and coarse estimates into accurate and complete prediction maps.



Specifically, we incorporate parallel axial attention \cite{ho2019axial} into the original reverse attention. To address the computational resource demand when calculating attention on high-dimensional features, we first obtain the feature map $F{F_i}$ from high-level features $\{ {F_i},i = 3,4,5\}$ through parallel computation aggregation along the horizontal and vertical axis to extract salient feature information:
\begin{equation}
    \begin{aligned}
    HorAtten(Q,K,V) & = VerAtten(Q,K,V) \\
              & = soft\max (\frac{{Q{K^T}}}{{\sqrt {{d_K}} }}),    
    \end{aligned}
\end{equation}
\begin{equation}
    F{F_i} = HorAtten({F_i}) + VerAtten({F_i}),
\end{equation}
where Q, K, V and ${d_K}$ denote query, key, value and the dimensions of key, respectively.

Then we inverse the feature map ${M_{i + 1}}$ from previous layer to obtain the reverse attention weight $r_i$:
\begin{equation}
{r_i} = 1 - \sigma (U({M_{i + 1}})),
\end{equation}where $\sigma $ denotes the Sigmoid function and $U$ denotes the upsampling operation.


Finally, we obtain the reverse attention features $F_i^r$ by multiplying axial aggregation feature $F{F_i}$ with reverse attention weight $r_i$ and then added back to the high-level feature ${F_i}$:
\begin{equation}
    {F_i^r} = F{F_i} \cdot {r_i} + {F_i},
\end{equation}


This operation efficiently directs the network to focus on delineation the predicted polyp area of deeper layer and surrounding tissues. 


\subsubsection{PA-BA}


As parallel axial reverse attention tends to focus more on the background regions of the image, there are still limitations in localizing the boundaries of polyps. To address this issue, we introduce Parallel Axial Boundary Attention to further improve the model's ability to precisely identify polyp edges. 

Similar with PA-RA, PA-BA first use axial attention \cite{ho2019axial} to generate the feature map $F{F_i}$ from high-level features $\{ {F_i},i = 3,4,5\}$. Subsequently, we use the feature map ${M_{i + 1}}$ from the previous layer and obtain the boundary attention weight $b_i$ according to ACSNet\cite{zhang2020adaptive}:
\begin{equation}
    {b_i} = 1 - \frac{{\left| {{\sigma (U({M_{i + 1}}))} - 0.5} \right|}}{{0.5}},
\end{equation}
where $\sigma $ denotes the Sigmoid function and $U$ denotes the upsampling operation.

The axial aggregated feature $F{F_i}$ is multiplied with boundary attention $b_i$ and added to the output feature map ${F_i}$ of the current layer to obtain the boundary attention features $F_i^b$:
\begin{equation}
    {F_i^b} = F{F_i} \cdot {b_i} + {F_i},
\end{equation}

Then the obtained attention features $F_i^r$ and $F_i^b$ from PA-RA and PA-BA are compressed with a convolution layer and then concatenated to obtain the parallel attention features $F_i^{rb}$ with enhanced boundary information.

Parallel Attention Module adaptively attend to the boundary information in three parallel high-level features. With axial attention, we achieved higher training efficiency. By integrating boundary attention, we addressed the deficiency of reverse attention to focus on boundary regions. PAM helps to improve the model's ability to identify boundaries and discriminate polyp region from background tissue, thus improving segmentation accuracy.



\subsection{Balancing Weight Module}

\begin{figure*}[!t]
\centering
\includegraphics[width=\textwidth]{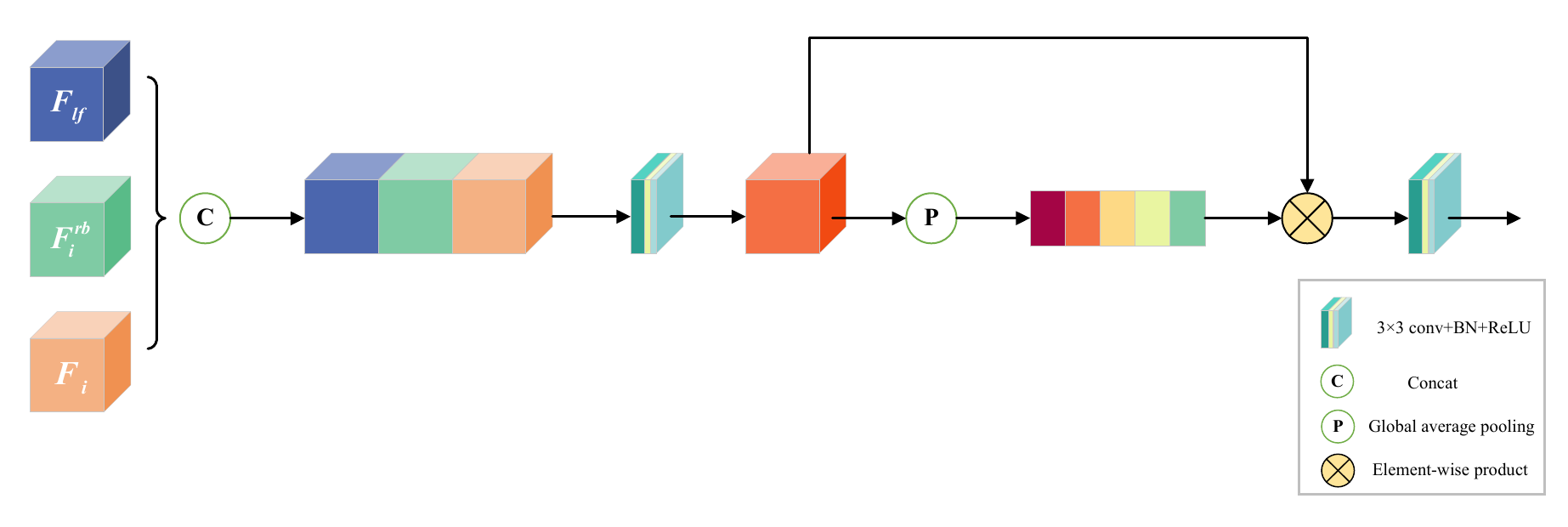}
\caption{Illustration of Balancing Weight Module (BWM).}
\label{fig_6}
\end{figure*}


In the bottom-up process of generating segmentation map, the upsampling operation can lead to blurriness. To address this problem, we introduce Balancing Weight Module to adaptively integrate low-level features ${F_{lf}}$, boundary attention $F_i^{rb}$ and high-level features $\{ {F_i},i = 3,4,5\}$, as shown in Figure \ref{fig_6}. This allows the network to concentrate on both local details and global context. 

To preserve local details in the decoding flow, it is crucial to incorporate low-level features. In contrast to "U-shaped" network that directly decoding low-level features, We employ the BWM module to incorporate low-level features as guidance for recognizing uncertain regions during the generation of segmentation maps for each deep layer. Specifically, the fused low-level features will first be filtered by CBAM \cite{woo2018cbam} (denoted as ${F_{clf}}$) to reduce noise and reinforce the boundary information from the low-level features. The parallel attention features $F_i^{rb}$ and the high-level features $\{ {F_i},i = 3,4,5\}$ from the current layer are resized to match the resolution of the ${F_{clf}}$. Then the three parts are concatenated to form the cascaded features.


The cascaded features are first processed by a $3 \times 3$ convolutional layer to reduce dimensionality. Then the compressed features are passed through a global average pooling layer followed by element-wise multiplication  to highlight contextual information for improving polyp segmentation accuracy. The segmentation map generated from the previous layer is added to the output of BWM to generate segmentation map of current layer.



The layer-by-layer generation of the segmentation map with BWM progressively refines and incorporates features from different levels, contributing to the accurate delineation of object boundaries in the final segmentation result.


\subsection{Loss Function}

We define the loss function as:
\begin{equation}
    {\cal L} = {\cal L}_{BCE}^\omega  + {\cal L}_{IoU}^\omega,
\end{equation}
where ${\cal L}_{BCE}^\omega $ \cite{wei2020f3net} and ${\cal L}_{IoU}^\omega $ \cite{wei2020f3net} denote the weighted binary cross entropy (BCE) loss and the weighted IoU loss, respectively. 

We found that in the training set, the sizes of polyps exhibit an obvious imbalance. Commonly used BCE loss can address the imbalance in positive sample segmentation. However, it primarily complements pixel-wise aspects at a microscopic level, which neglects the global structure within the image. This leads to certain deficiencies in learning challenging samples. Therefore, we supplement the IoU loss to constrain from the global aspect. 
Compared with standard BCE loss and IoU loss, ${\cal L}_{BCE}^\omega$ and ${\cal L}_{IoU}^\omega$ can emphasize the significance of small objects and boundary information by assigning larger weights to them. The effectiveness of these two methods has been confirmed in various studies \cite{ferrari_reverse_2018} \cite{bai2022end}.

We employ deep supervision for the three high-level feature maps $\{ {F_i},i = 3,4,5\}$ and the initial guidance map $F_{fuse}$.
These maps are upsampled (denoted as $F_{fuse}^{up}$ , $F_3^{up}$ , $F_4^{up}$ , $F_5^{up}$) to match the same size of the ground truth $G$. 

Finally, the whole segmentation framework can be trained in an end-to-end manner with total loss function:
\begin{equation}
\begin{aligned}
{{\cal L}_{total}} = &{\cal L}(G,F_{fuse}^{up}) + {\cal L}(G,F_3^{up}) \\
                  & + {\cal L}(G,F_4^{up}) + {\cal L}(G,F_5^{up}).
\end{aligned}
\end{equation}

\section{Experiments}

\subsection{Datasets}
We utilize the same dataset, and training and testing splits with PraNet for polyp segmentation.
Specifically, the dataset consists of 900 images from Kvasir-SEG and 550 images from CVC-ClinicDB, totaling 1450 samples. They are randomly divided into 80\% for training, 10\% for validation, and 10\% for testing.
We evaluate the performance of the proposed method on five benchmark datasets.
Detailed information of benchmarks is described below.

(1) CVC-T (CVC-300) \cite{vazquez_benchmark_2017}: this dataset contains 60 samples from 44 colonoscopy sequences from 36 patients, where all images are $500 \times 574$ in size. All of them are used for testing.

(2) CVC-ClinicDB (CVC-612) \cite{bernal_wm-dova_2015}: this dataset contains 612 images extracted from 29 different endoscopic video clips with very similar polyp targets. The size of all the images is $384 \times 288$. 62 images of this dataset are used for test and the rest of the images are used for training.

(3) CVC-ColonDB \cite{tajbakhsh_automated_2016}: this dataset contains 380 images from 15 different colonoscopy sequences with an image size of $500 \times 570$, all of which are used for testing.

(4) ETIS-LaribPolypDB \cite{silva_toward_2014}: this dataset contains 196 images collected from 34 colonoscopy videos. The size of all the images is $1225 \times 966$. This dataset is currently the most difficult in the field of polyp segmentation. Due to the smaller size and concealed locations of the polyps in this dataset, detection becomes challenging.

(5) Kvasir \cite{jha_kvasir-seg_2020}: this dataset contains 1000 images of polyps with sizes ranging from $332 \times 487$ to $1920 \times 1072$. The images exhibit variations in the size and shape of the polyps. For training purposes, 900 images from this dataset are utilized, while 100 images are reserved for testing.

\subsection{Evaluation Metrics}

To comprehensively assess our model, we employ the following metrics: 

(1) MeanDice and MeanIoU are used to measure the similarity between predicted segmentation results and ground truth. Larger values denote higher similarity between the predicted segmentation results and the ground truth.

(2) $F_\beta ^\omega $ ($\omega Fm$) \cite{margolin_how_2014} intuitively generalized the F-measure \cite{achanta_frequency-tuned_2009} by calculating the accuracy and recall alternately, assigning different weights to various errors in different locations of neighborhood information. This approach is used to correct the "equal-importance defect" in Dice \cite{borji2015salient}:
\begin{equation}
F_\beta ^\omega  = \frac{{(1 + {\beta ^2}){{\mathop{\rm P}\nolimits} ^\omega } \times {R^\omega }}}{{{\beta ^2}{{\mathop{\rm P}\nolimits} ^\omega } + {R^\omega }}},
\end{equation}
where $P$ denotes precision and $R$ denotes recall. $\beta$ and $\omega$ are used to adjust the relative importance of precision and recall, respectively.

(3) S-measure (${S_m}$) \cite{fan_structure-measure_2017} is used to evaluate the structural similarity between the prediction segmentation results and the ground truth:
\begin{equation}
S = \alpha  \times {S_0} + (1 - \alpha ) \times {S_r},
\end{equation}
where ${S_0}$ and ${S_r}$ denote the object-aware and region-aware structural similarity, respectively. $\alpha$ is set to 0.5 in our experiments.

(4) E-measure ($E_\phi ^{\max }$) \cite{fan2018enhanced} is an enhanced alignment method measured from the perspective of global average and local pixel matching with the ground truth: 
\begin{equation}
{Q_s} = \frac{1}{{W \times H}}\sum\nolimits_{i = 1}^W {\sum\nolimits_{j = 1}^H {{\phi _s}(i,j)} },
\end{equation}
where ${\phi _s}$ denotes the enhanced alignment matrix, which reflects the correlation between the prediction segmentation results and the ground truth.


(5) MAE \cite{perazzi_saliency_2012} is the average pixel-wise absolute error, which is used to evaluate the pixel-level accuracy between the ground truth and the predicted segmentation results:
\begin{equation}
MAE = \frac{1}{{W \times H}}\sum\nolimits_{i = 1}^W {\sum\nolimits_{j = 1}^H {\left| {G(i,j) - S(i,j)} \right|} },
\end{equation}
where $G$ denotes the ground truth and $S$ denotes the predicted segmentation results.

\begin{table*}[htbp]
  \centering
  \caption{Quantitative comparison on Kvasir and CVC-ClinicDB datasets.
  \textbf{Bold} and \uline{underlined} represent the top-2 results respectively.}
  \begin{tabular*}{\linewidth}{m{1cm}<{\centering}|m{2.5cm}<{\centering}m{1.9cm}<{\centering}m{1.9cm}<{\centering}m{1.9cm}<{\centering}m{1.9cm}<{\centering}m{1.9cm}<{\centering}m{1.5cm}<{\centering}}

    \hline
    \multicolumn{1}{c|}{\textbf{Dataset}}
        & {\textbf{Methods}} & {\textbf{mDice}} & {\textbf{mIoU}} & {\textbf{$\bm{F_\beta^\omega}$}} & {\textbf{$\bm{S_m}$}} & {\textbf{$\bm{E_\phi^{\max}}$}} & {\textbf{MAE}} \\
    \hline

    \hline
    \multicolumn{1}{c|}{\multirow{18}{*}{Kvasir}}
          & UNet \cite{ronneberger_u-net_2015}  & 0.870 & 0.806 & 0.855 & 0.885 & 0.924 & 0.039 \\
          & UNet++ \cite{zhou2018unet++} & 0.866 & 0.804 & 0.851 & 0.882 & 0.918 & 0.038 \\
          & SegNet \cite{wickstrom2020uncertainty} & 0.880 & 0.812 & 0.856 & 0.889 & 0.934 & 0.035 \\
          & ResUNet \cite{diakogiannis2020resunet} & 0.774 & 0.675 & 0.744 & 0.809 & 0.873 & 0.062 \\
          & ResUNet++  \cite{jha_resunet_2019} & 0.867 & 0.798 & 0.848 & 0.882 & 0.923 & 0.041 \\
          & U2Net \cite{qin2020u2} & 0.896 & 0.843 & 0.889 & 0.909 & 0.937 & 0.031 \\
          & PraNet \cite{fan_pranet_2020} & \uline{0.899} & \textbf{0.850} & \uline{0.891} & \uline{0.912} & \uline{0.944} & 0.029 \\
          & C2FNet \cite{sun2021context} & 0.897 & \uline{0.847} & 0.891 & 0.910 & 0.939 & 0.029 \\
          & MSNet \cite{zhao_automatic_2021} & 0.891 & 0.841 & 0.880 & 0.907 & 0.942 & \uline{0.027} \\
          & CaraNet \cite{lou_caranet_2023} & 0.876 & 0.820 & 0.869 & 0.894 & 0.925 & 0.037 \\
          & DoubleU-Net \cite{li2023erdunet} & 0.894 & 0.833 & 0.883 & 0.898 & 0.940 & 0.038 \\
          & FCBFormer \cite{sanderson2022fcn} & 0.893 & 0.835 & 0.884 & 0.901 & 0.942 & 0.029 \\
          & GMSRF-Net \cite{srivastava2022gmsrf} & 0.858 & 0.787 & 0.847 & 0.878 & 0.917 & 0.044 \\
          & HarDNet-MSEG \cite{huang2021hardnet} & 0.882 & 0.821 & 0.870 & 0.895 & 0.927 & 0.036 \\
          & Polyp-PVT \cite{dong2021polyp} & 0.854 & 0.783 & 0.842 & 0.875 & 0.921 & 0.037 \\
          & TransFuse-S \cite{zhang2021transfuse} & 0.859 & 0.783 & 0.841 & 0.877 & 0.926 & 0.038 \\
          & TransFuse-L \cite{zhang2021transfuse} & 0.863 & 0.787 & 0.846 & 0.880 & 0.929 & 0.036 \\
          & \textbf{Ours} & \textbf{0.903} & 0.846 & \textbf{0.902} & \textbf{0.915} & \textbf{0.947} & \textbf{0.025} \\
          
    \hline
    \multicolumn{1}{c|}{\multirow{18}{*}{CVC-ClinicDB}} 
          & UNet \cite{ronneberger_u-net_2015}  & 0.889 & 0.831 & 0.889 & 0.920 & 0.950 & 0.014 \\
          & UNet++ \cite{zhou2018unet++} & 0.900 & 0.849 & 0.901 & 0.926 & 0.960 & 0.012 \\
          & SegNet \cite{wickstrom2020uncertainty} & 0.852 & 0.790 & 0.844 & 0.894 & 0.924 & 0.017 \\
          & ResUNet \cite{diakogiannis2020resunet} & 0.801 & 0.721 & 0.798 & 0.859 & 0.913 & 0.027 \\
          & ResUNet++ \cite{jha_resunet_2019} & \uline{0.910} & 0.857 & \uline{0.912} & 0.932 & \uline{0.968} & 0.012 \\
          & U2Net \cite{qin2020u2} & 0.906 & 0.859 & 0.903 & 0.929 & 0.962 & 0.015 \\
          & PraNet \cite{fan_pranet_2020} & 0.903 & \uline{0.859} & 0.902 & \uline{0.935} & 0.961 & 0.008 \\
          & C2FNet \cite{sun2021context} & 0.902 & 0.857 & 0.899 & 0.932 & 0.967 & 0.009 \\
          & MSNet \cite{zhao_automatic_2021} & 0.900 & 0.857 & 0.899 & 0.933 & 0.965 & \uline{0.008} \\
          & CaraNet \cite{lou_caranet_2023} & 0.884 & 0.829 & 0.884 & 0.914 & 0.951 & 0.015 \\
          & DoubleU-Net \cite{jha2020doubleu} & 0.878 & 0.822 & 0.871 & 0.910 & 0.947 & 0.017 \\
          & FCBFormer \cite{sanderson2022fcn} & 0.901 & 0.855 & 0.901 & 0.927 & 0.953 & 0.016 \\
          & GMSRF-Net \cite{srivastava2022gmsrf} & 0.847 & 0.786 & 0.849 & 0.892 & 0.935 & 0.019 \\
          & HarDNet-MSEG \cite{huang2021hardnet} & 0.895 & 0.845 & 0.891 & 0.921 & 0.961 & 0.009 \\
          & Polyp-PVT \cite{dong2021polyp} & 0.855 & 0.787 & 0.854 & 0.891 & 0.953 & 0.019 \\
          & TransFuse-S \cite{zhang2021transfuse} & 0.848 & 0.770 & 0.840 & 0.886 & 0.942 & 0.018 \\
          & TransFuse-L \cite{zhang2021transfuse} & 0.853 & 0.780 & 0.849 & 0.894 & 0.947 & 0.018 \\
          & \textbf{Ours} & \textbf{0.918} & \textbf{0.869} & \textbf{0.927} & \textbf{0.935} & \textbf{0.971} & \textbf{0.008} \\
        \hline
    \end{tabular*}%
  \label{tab:1}%
\end{table*}%

\subsection{Implementation details}

We use Res2Net-50 pretrained on ImageNet as the backbone for feature extraction. We use Adam optimizer with the initial learning rate of 1e-5, and weight decay of 1e-5 to optimize our model. The batch size is set to 16.

In training, all images are resized to 352$\times$352. To ensure better model convergence, the total number of training epochs is set to 300.
To enhance model stability in later training stages, a Poly learning rate decay strategy is employed, represented as $lr = base\_lr \times {(1 - \frac{{epoch}}{{total\_epoch}})^{power}}$ , where the power is set to 0.9. Our proposed network is implemented using PyTorch and trained on a Tesla V100 with 32GB of memory. During training, extensive data augmentation is applied on-the-fly to improve the generalization, including
random scaling and cropping, flipping, Gaussian noise, contrast, brightness and sharpness variations. For the ground truth, dilation and erosion operations are applied with kernels vary from 2 to 5. 




\subsection{Comparison with State-of-the-art Methods}

In order to validate the effectiveness of the proposed model, we compare our method to state-of-the-art segmentation methods: UNet \cite{ronneberger_u-net_2015}, UNet++ \cite{zhou2018unet++}, SegNet \cite{wickstrom2020uncertainty}, ResUNet \cite{diakogiannis2020resunet}, ResUNet++ \cite{jha_resunet_2019}, U2Net \cite{qin2020u2}, PraNet \cite{fan_pranet_2020}, C2FNet \cite{sun2021context}, MSNet \cite{zhao_automatic_2021}, CaraNet \cite{lou_caranet_2023}, DoubleU-Net\cite{jha2020doubleu}, FCBFormer\cite{sanderson2022fcn}, GMSRF-Net\cite{srivastava2022gmsrf}, HarDNet-MSEG\cite{huang2021hardnet}, Polyp-PVT\cite{dong2021polyp}, TransFuse\cite{zhang2021transfuse}. To ensure fair comparison, we use official implementations of these comparison models and apply the same data augmentation strategy and dataset splits. All experiments are conducted in the same environment. We report quantitative comparisons on test sets of Kvasir and CVC-ClinicDB in Table \ref{tab:1} to validate our model's learning ability, and comparisons on unseen datasets CVC-300, CVC-ColonDB and ETIS in Table \ref{tab:2} to verify the model's generalizability.

\subsubsection{Quantitative Comparison}


As shown in Table \ref{tab:1}, our model outperforms all classic baselines on CVC-ClinicDB dataset. 
On Kvasir dataset, our proposed method achieves the best performance across five metrics and comparable results on meanIOU. This indicates that our method have learned accurate local detail and global context to correctly segment polyps.  

\begin{table*}[htbp]
  \centering
  \caption{Genearlization comparison on CVC-300, CVC-ColonDB and ETIS datasets. \textbf{Bold} and \uline{underlined} represent the top-2 results respectively.}
 
  \begin{tabular*}{\linewidth}{m{1cm}<{\centering}|m{2.5cm}<{\centering}m{1.9cm}<{\centering}m{1.9cm}<{\centering}m{1.9cm}<{\centering}m{1.9cm}<{\centering}m{1.9cm}<{\centering}m{1.5cm}<{\centering}}

    \hline
    \multicolumn{1}{c|}{\textbf{Dataset}}
        & {\textbf{Methods}} & {\textbf{mDice}} & {\textbf{mIoU}} & {\textbf{$\bm{F_\beta^\omega}$}} & {\textbf{$\bm{S_m}$}} & {\textbf{$\bm{E_\phi^{\max}}$}} & {\textbf{MAE}} \\
    \hline

    \hline
    \multicolumn{1}{c|}{\multirow{18}{*}{CVC-300}} 
    
          & UNet \cite{ronneberger_u-net_2015}  & 0.823 & 0.742 & 0.803 & 0.883 & 0.928 & 0.013 \\
          & UNet++ \cite{zhou2018unet++} & 0.839 & 0.754 & 0.812 & 0.891 & 0.942 & 0.013 \\
          & SegNet \cite{wickstrom2020uncertainty} & 0.760 & 0.670 & 0.710 & 0.839 & 0.889 & 0.018 \\
          & ResUNet \cite{diakogiannis2020resunet} & 0.517 & 0.399 & 0.505 & 0.697 & 0.756 & 0.026 \\
          & ResUNet++ \cite{jha_resunet_2019} & 0.815 & 0.719 & 0.776 & 0.877 & 0.930 & 0.017 \\
          & U2Net \cite{qin2020u2} & 0.809 & 0.722 & 0.767 & 0.879 & 0.902 & 0.019 \\
          & PraNet \cite{fan_pranet_2020} & 0.864 & 0.785 & 0.837 & 0.902 & 0.948 & 0.010 \\
          & C2FNet \cite{sun2021context} & 0.845 & 0.772 & 0.821 & 0.902 & 0.937 & 0.013 \\
          & MSNet \cite{zhao_automatic_2021} & 0.838 & 0.771 & 0.818 & 0.895 & 0.940 & 0.011 \\
          & CaraNet \cite{lou_caranet_2023} & 0.854 & 0.787 & 0.837 & 0.907 & 0.942 & 0.010 \\
          & DoubleU-Net \cite{jha2020doubleu} & 0.793 & 0.712 & 0.751 & 0.868 & 0.891 & 0.021 \\
          & FCBFormer \cite{sanderson2022fcn} & \uline{0.889} & \uline{0.817} & \uline{0.865} & \uline{0.925} & \uline{0.964} & \uline{0.008} \\
          & GMSRF-Net \cite{srivastava2022gmsrf} & 0.769 & 0.675 & 0.727 & 0.850 & 0.891 & 0.022 \\
          & HarDNet-MSEG \cite{srivastava2022gmsrf} & 0.833 & 0.766 & 0.811 & 0.900 & 0.937 & 0.016 \\
          & Polyp-PVT \cite{shi2022polyp} & 0.841 & 0.748 & 0.820 & 0.892 & 0.949 & 0.011 \\
          & TransFuse-S \cite{zhang2021transfuse} & 0.809 & 0.717 & 0.766 & 0.872 & 0.922 & 0.014 \\
          & TransFuse-L \cite{zhang2021transfuse} & 0.834 & 0.741 & 0.798 & 0.890 & 0.940 & 0.011 \\
          & \textbf{Ours} & \textbf{0.907} & \textbf{0.842} & \textbf{0.893} & \textbf{0.935} & \textbf{0.980} & \textbf{0.005} \\
    \hline
    \multicolumn{1}{c|}{\multirow{18}{*}{CVC-ColonDB}} 
          & UNet \cite{ronneberger_u-net_2015} & 0.657 & 0.566 & 0.633 & 0.768 & 0.825 & 0.050 \\
          & UNet++ \cite{zhou2018unet++} & 0.652 & 0.562 & 0.629 & 0.769 & 0.814 & 0.052 \\
          & SegNet \cite{wickstrom2020uncertainty} & 0.677 & 0.589 & 0.643 & 0.777 & 0.840 & 0.050 \\
          & ResUNet \cite{diakogiannis2020resunet} & 0.506 & 0.397 & 0.471 & 0.676 & 0.766 & 0.061 \\
          & ResUNet++ \cite{jha_resunet_2019} & 0.695 & 0.604 & 0.662 & 0.796 & 0.852 & 0.050 \\
          & U2Net \cite{qin2020u2} & 0.734 & 0.655 & 0.709 & 0.817 & 0.868 & 0.045 \\
          & PraNet \cite{fan_pranet_2020} & 0.748 & \uline{0.676} & 0.736 & 0.828 & 0.870 & 0.043 \\
          & C2FNet \cite{sun2021context} & 0.740 & 0.664 & 0.726 & 0.827 & 0.859 & 0.039 \\
          & MSNet \cite{zhao_automatic_2021} & 0.746 & 0.665 & 0.732 & 0.825 & 0.866 & 0.039 \\
          & CaraNet \cite{lou_caranet_2023} & 0.744 & 0.665 & \uline{0.740} & 0.827 & 0.867 & 0.038 \\
          & DoubleU-Net \cite{jha2020doubleu} & 0.709 & 0.630 & 0.691 & 0.805 & 0.848 & 0.044 \\
          & FCBFormer \cite{sanderson2022fcn} & \uline{0.754} & 0.671 & 0.738 & \uline{0.830} & \uline{0.889} & \uline{0.037} \\
          & GMSRF-Net \cite{srivastava2022gmsrf} & 0.675 & 0.585 & 0.646 & 0.781 & 0.828 & 0.052 \\
          & HarDNet-MSEG \cite{huang2021hardnet} & 0.722 & 0.650 & 0.711 & 0.814 & 0.858 & 0.042 \\
          & Polyp-PVT \cite{dong2021polyp} & 0.666 & 0.564 & 0.656 & 0.773 & 0.829 & 0.044 \\
          & TransFuse-S \cite{zhang2021transfuse} & 0.689 & 0.597 & 0.660 & 0.788 & 0.862 & 0.045 \\
          & TransFuse-L \cite{zhang2021transfuse} & 0.711 & 0.617 & 0.683 & 0.798 & 0.874 & 0.044 \\
          & \textbf{Ours}  & \textbf{0.762} & \textbf{0.690} & \textbf{0.754} & \textbf{0.838} & \textbf{0.892} & \textbf{0.036} \\
    \hline

    \multicolumn{1}{c|}{\multirow{18}{*}{ETIS}}        
          & UNet \cite{ronneberger_u-net_2015} & 0.542 & 0.476 & 0.514 & 0.735 & 0.757 & 0.030 \\
          & UNet++ \cite{zhou2018unet++} & 0.533 & 0.470 & 0.504 & 0.731 & 0.749 & 0.029 \\
          & SegNet \cite{wickstrom2020uncertainty} & 0.550 & 0.476 & 0.509 & 0.733 & 0.768 & 0.034 \\
          & ResUNet \cite{diakogiannis2020resunet} & 0.356 & 0.275 & 0.328 & 0.633 & 0.644 & 0.042 \\
          & ResUNet++ \cite{jha_resunet_2019} & 0.470 & 0.393 & 0.420 & 0.677 & 0.734 & 0.054 \\
          & U2Net \cite{qin2020u2} & 0.655 & 0.577 & 0.606 & 0.793 & 0.840 & 0.026 \\
          & PraNet \cite{fan_pranet_2020} & 0.557 & 0.491 & 0.524 & 0.739 & 0.767 & 0.054 \\
          & C2FNet \cite{sun2021context} & 0.624 & 0.560 & 0.596 & 0.780 & 0.813 & 0.029 \\
          & MSNet \cite{zhao_automatic_2021} & 0.582 & 0.513 & 0.550 & 0.747 & 0.791 & 0.036 \\
          & CaraNet \cite{lou_caranet_2023} & 0.694 & 0.618 & 0.668 & 0.813 & 0.868 & 0.019 \\
          & DoubleU-Net \cite{jha2020doubleu} & 0.656 & 0.579 & 0.618 & 0.790 & 0.854 & 0.024 \\
          & FCBFormer \cite{sanderson2022fcn} & \uline{0.708} & \uline{0.633} & \uline{0.673} & \uline{0.829} & \uline{0.873} & 0.020 \\
          & GMSRF-Net \cite{srivastava2022gmsrf} & 0.528 & 0.444 & 0.499 & 0.725 & 0.783 & 0.031 \\
          & HarDNet-MSEG \cite{huang2021hardnet} & 0.590 & 0.519 & 0.553 & 0.754 & 0.807 & 0.045 \\
          & Polyp-PVT \cite{dong2021polyp} & 0.654 & 0.566 & 0.631 & 0.791 & 0.863 & \uline{0.017} \\
          & TransFuse-S \cite{zhang2021transfuse} & 0.604 & 0.502 & 0.547 & 0.743 & 0.821 & 0.028 \\
          & TransFuse-L \cite{zhang2021transfuse} & 0.551 & 0.462 & 0.507 & 0.724 & 0.816 & 0.026 \\
          & \textbf{Ours}  & \textbf{0.764} & \textbf{0.686} & \textbf{0.739} & \textbf{0.855} & \textbf{0.900} & \textbf{0.012} \\
    \hline
    \end{tabular*}%

    \label{tab:2}%
\end{table*}%

\textbf{Generalization Capability.} From Table \ref{tab:2} we can see that on three unseen datasets, our model still surpasses other approaches across all metrics, demonstrating strong generalizability of our method. 
On the most challenging ETIS dataset, one notable findings is that all SOTA methods experienced a significant decline in performance (PraNet dropped 38.3\% on meanDice), while our exhibited a much smaller decrease (16.8\% on meanDice). By properly integrating multi-scale features, our model can well handle the scale variations of polyps. Besides, the boundary details is effectively enhanced and refined to delineate the polyps and surrounding tissues, thereby well generalized to unseen colonoscopy.


\subsubsection{Qualitative Comparison}

\begin{figure*}[!t]  
\centering 
\includegraphics[width = \textwidth]{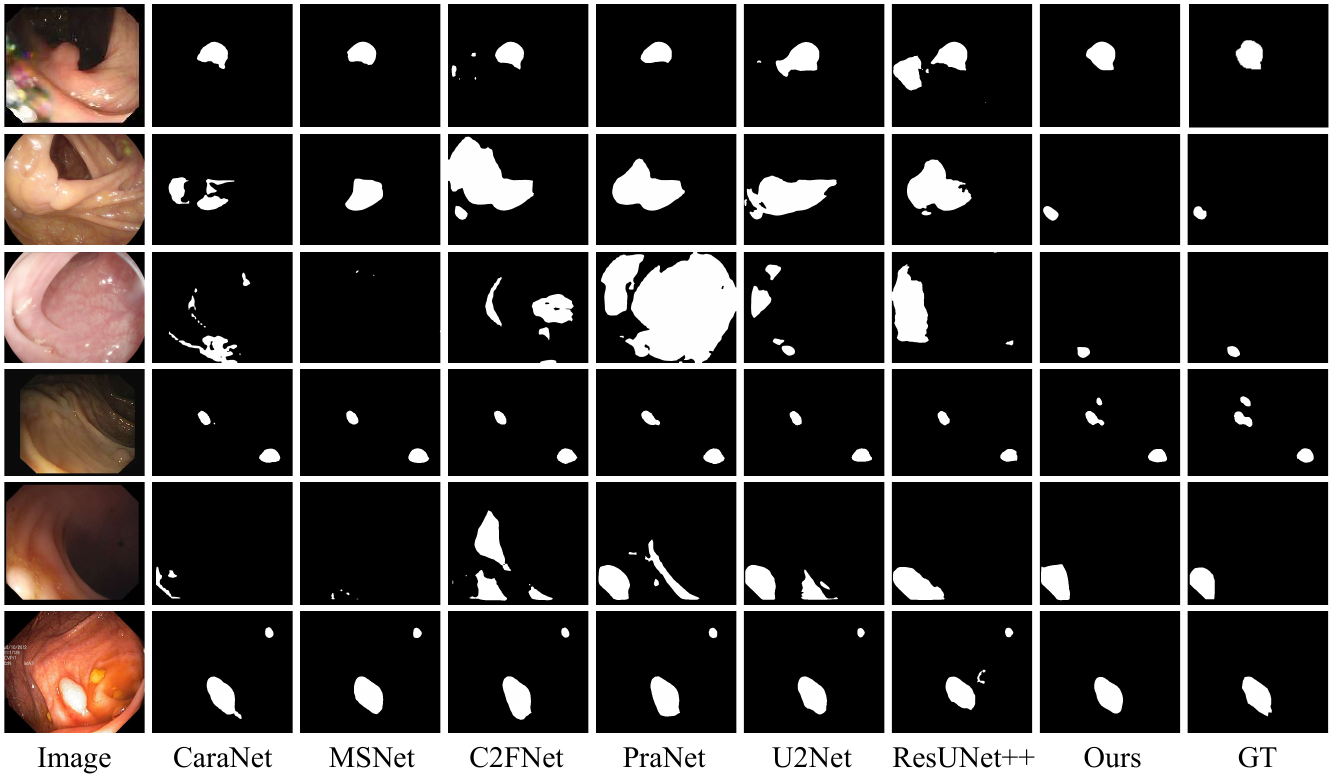} 
\caption{Qualitative comparison with baseline segmentation methods.}
\label{fig_7}
\end{figure*}

In Figure \ref{fig_7}, we provide visual comparison of the segmentation results. As illustrated, our model significantly improved the accuracy and clarity of polyp segmentation in handling different situations.
In the second and third rows, the polyp is located in a concealed position. Almost all other comparative methods fail to predict accurate segmentation, while our model can segment the polyps almost completely accurately. As shown in the fourth row, with low lighting condition and different sizes of polyps, our method managed to segment all the polyps. The visual results further proves that our model can better cope with challenges posed by varying lighting conditions, sizes, shapes, and other factors. 

\section{Ablation Study}

\begin{figure}  
\centering 
\includegraphics[width = 0.4\textwidth]{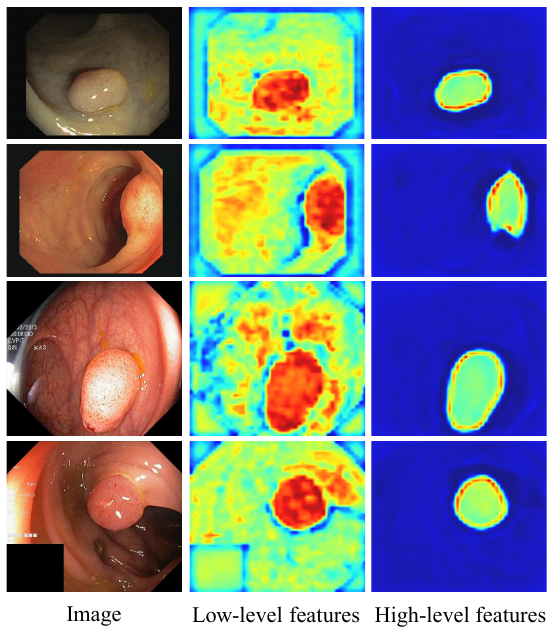} 
\caption{Visualization of low-level and high-level feature attention $ g_c $ and $ h_c $ in Eq. 7.}
\label{fig_9}
\end{figure}

\begin{figure*}  
\centering 
\includegraphics[width = \textwidth]{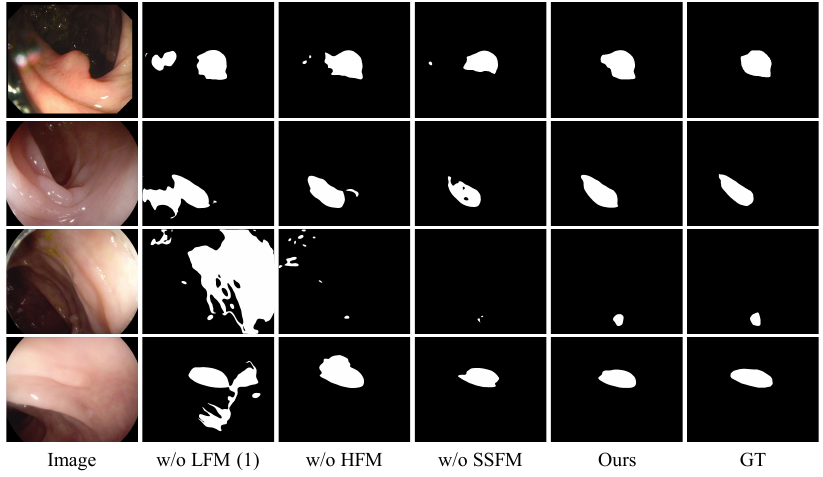} 
\caption{Visual comparisons about ablation study.}
\label{fig_8}
\end{figure*}

We conduct a series of ablation studies to verify the effectiveness of the key components in our framework. We report the ablation study results in Table \ref{tab:4}, Table \ref{tab:5} and Table \ref{tab:6} on Kvasir, CVC-ClinicDB, CVC-300, CVC-ColonDB and ETIS datasets. Figure \ref{fig_8} and \ref{fig_9} illustrate visual comparisons of the ablation studies across these five datasets.


\begin{table}[htbp]
  \centering
  \caption{The settings of different ablation studies for SSFM.}
  \begin{tabular}{m{2cm}<{\centering}|m{1.2cm}<{\centering}m{1.2cm}<{\centering}m{1.2cm}<{\centering}}
    \hline
    \multicolumn{1}{c|}{\textbf{Settings}} & {\textbf{LFM}} & {\textbf{HFM}} & {\textbf{SSFM}}  \\
    \hline
    w/o LFM (1) & \ding{55} & \ding{51} & \ding{55} \\
    w/o HFM & \ding{51} & \ding{55} & \ding{55} \\
    w/o SSFM & \ding{51} & \ding{51} & \ding{55} \\
    Ours & \ding{51} & \ding{51} & \ding{51} \\
    \hline
  \end{tabular}%
  \label{tab:3}%
\end{table}%

\textbf{Effectiveness of Selectively Shared Fusion Module.} To verify the effects of SSFM, we compare our model with three variants, as shown in Table \ref{tab:3}: (1) \emph{w/o LFM(1)}: we remove the low-level feature input of SSIM. The initial guidance map is generated with only integrated high-level features. (2) \emph{w/o HFM}: we remove the high-level feature input of SSIM. The initial guidance map is generated with only integrated low-level features. (3) \emph{w/o SSFM}: we directly fuse the low-level feature and high-level feature using addition operation.

As shown in Table \ref{tab:4} and Figure \ref{fig_8}, when initial guidance map is generated with only low-level features (w/o HFM) or only high-level features (w/o LFM(1)), the segmentation accuracy drops. Among them, high-level features contribute more to performance. Also, the feature fusion by directly adding low-level and high-level features(w/o SSFM) failed to improve model performance. Instead, it resulted in a decrease in model performance, indicating the importance of an appropriate feature fusion method. 

In contrast, SSFM noticeably improved the segmentation accuracy. By enforcing information sharing and adaptive selection between different level of features, SSFM enhances the model's ability to capture diverse information. Specifically, on the challenging ETIS dataset, SSFM can effectively adapt to the scale variations of polyps, greatly improved the generalizability of our method.


In Figure \ref{fig_9}, we visualize the attention $g_c$ and $h_c$ in adaptive selection of SSFM. We can see that adaptive selection would concentrate on discriminating the boundary between a polyp and its surrounding tissues when attend to low-level features. Conversely, when attend to high-level features, accurate location of polyp regions are emphasized. Figure \ref{fig_9} explicitly indicated that features of different scales contribute diverse information to the model, and SSFM effectively integrates these information, therefore enhancing the model's performance.






\begin{table*}[htbp]
  \centering
  \caption{Ablation study for selectively shared feature fusion on Kvasir, CVC-ClinicDB, CVC-300, CVC-ColonDB and ETIS datasets.}
  \begin{tabular*}{\linewidth}{m{1cm}<{\centering}|m{2.5cm}<{\centering}m{1.9cm}<{\centering}m{1.9cm}<{\centering}m{1.9cm}<{\centering}m{1.9cm}<{\centering}m{1.9cm}<{\centering}m{1.5cm}<{\centering}}
    \hline
    \multicolumn{1}{c|}{\textbf{Dataset}} 
    & {\textbf{Settings}} & {\textbf{meanDice}} & {\textbf{meanIoU}} & {\textbf{$\bm{F_\beta^\omega}$}} & {\textbf{$\bm{S_m}$}} & {\textbf{$\bm{E_\phi^{\max}}$}} & {\textbf{MAE}} \\
    \hline
    \multicolumn{1}{c|}{\multirow{4}[2]{*}{Kvasir}}
          & w/o LFM (1) & 0.887  & 0.830  & 0.878  & 0.898  & 0.935  & 0.035  \\
          & w/o HFM & 0.839  & 0.768  & 0.811  & 0.862  & 0.899  & 0.052  \\
          & w/o SSFM & 0.891  & 0.834  & 0.888  & 0.907  & 0.938  & 0.028  \\
          & \textbf{Ours} & \textbf{0.903 } & \textbf{0.846 } & \textbf{0.902 } & \textbf{0.915 } & \textbf{0.947 } & \textbf{0.025 } \\
    \hline
    \multicolumn{1}{c|}{\multirow{4}[2]{*}{CVC-ClinicDB}}
          & w/o LFM (1) & 0.888  & 0.835  & 0.885  & 0.919  & 0.950  & 0.015  \\
          & w/o HFM & 0.845  & 0.776  & 0.833  & 0.889  & 0.925  & 0.018  \\
          & w/o SSFM & 0.880  & 0.823  & 0.894  & 0.910  & 0.945  & 0.009  \\
          & \textbf{Ours} & \textbf{0.918 } & \textbf{0.869 } & \textbf{0.927 } & \textbf{0.935 } & \textbf{0.971 } & \textbf{0.008 } \\
    \hline
    \multicolumn{1}{c|}{\multirow{4}[2]{*}{CVC-300}}
          & w/o LFM (1) & 0.860  & 0.786  & 0.837  & 0.906  & 0.939  & 0.010  \\
          & w/o HFM & 0.874  & 0.805  & 0.855  & 0.913  & 0.952  & 0.009  \\
          & w/o SSFM & 0.876  & 0.798  & 0.882  & 0.908  & 0.961  & 0.007  \\
          & \textbf{Ours} & \textbf{0.907 } & \textbf{0.842 } & \textbf{0.893 } & \textbf{0.935 } & \textbf{0.980 } & \textbf{0.005 } \\
    \hline
    \multicolumn{1}{c|}{\multirow{4}[2]{*}{CVC-ColonDB}}
          & w/o LFM (1) & 0.734  & 0.656  & 0.714  & 0.814  & 0.839  & 0.056  \\
          & w/o HFM & 0.708  & 0.625  & 0.689  & 0.804  & 0.841  & 0.044  \\
          & w/o SSFM & 0.700  & 0.616  & 0.706  & 0.799  & 0.829  & 0.040  \\
          & \textbf{Ours} & \textbf{0.762 } & \textbf{0.690 } & \textbf{0.754 } & \textbf{0.838 } & \textbf{0.892 } & \textbf{0.036 } \\
    \hline
    \multicolumn{1}{c|}{\multirow{4}[2]{*}{ETIS}}
          & w/o LFM (1) & 0.582  & 0.509  & 0.532  & 0.725  & 0.714  & 0.108  \\
          & w/o HFM & 0.701  & 0.620  & 0.665  & 0.813  & 0.871  & 0.027  \\
          & w/o SSFM & 0.600  & 0.534  & 0.592  & 0.764  & 0.798  & 0.014  \\
          & \textbf{Ours} & \textbf{0.764 } & \textbf{0.686 } & \textbf{0.739 } & \textbf{0.855 } & \textbf{0.900 } & \textbf{0.012 } \\
    \hline
    \end{tabular*}%
  \label{tab:4}%
\end{table*}%

\textbf{Effectiveness of integrating low-level features.} In our method, low-level features serve two primary purposes. Firstly, the fused low-level features are used in SSFM module to generate the initial guidance map. Secondly, the boundary information contained in the low-level features contributes to continuously refinement of segmentation maps with BWM module. To validate the effectiveness of integrating low-level features, we compare our model with two variants: w/o LFM (1) and w/o LFM(2), where remove the integrated low-level feature from Balancing Weight Module. 



As shown in Table \ref{tab:4}, \ref{tab:5} and \ref{tab:6}, low-level features have been proved worthy for improving segmentation accuracy. The visual comparison results in Figure \ref{fig_8} and Figure \ref{fig_9} further indicate that the integrated low-level features is crucial for initial guidance map generation and the enhancement of boundary information, significantly improving the clarity of segmentation results.

\begin{table*}[htbp]
  \centering
  \caption{Ablation study on the Kvasir and CVC-ClinicDB datasets.}
    \begin{tabular*}{\linewidth}{m{1cm}<{\centering}|m{2.5cm}<{\centering}m{1.9cm}<{\centering}m{1.9cm}<{\centering}m{1.9cm}<{\centering}m{1.9cm}<{\centering}m{1.9cm}<{\centering}m{1.5cm}<{\centering}}
      
    \hline
    \multicolumn{1}{c|}{\textbf{Dataset}}
          & {\textbf{Settings}} & {\textbf{meanDice}} & {\textbf{meanIoU}} & {\textbf{$\bm{F_\beta^\omega}$}} & {\textbf{$\bm{S_m}$}} & {\textbf{$\bm{E_\phi^{\max}}$}} & {\textbf{MAE}} \\
    \hline

    \multicolumn{1}{c|}{\multirow{7}{*}{Kvasir}} 
          & w/o LFM (2) & 0.878  & 0.820  & 0.863  & 0.893  & 0.924  & 0.037  \\
          & w/o PAM & 0.856  & 0.789  & 0.862  & 0.885  & 0.917  & 0.039  \\
          & PAM (PA-RA) & 0.898  & \textbf{0.847 } & 0.892  & 0.911  & 0.941  & 0.029  \\
          & PAM (PA-BA) & 0.887  & 0.830  & 0.880  & 0.899  & 0.930  & 0.036  \\
          & w/o BWM & 0.893  & 0.839  & 0.892  & 0.909  & 0.944  & 0.029  \\
          & \textbf{Ours} & \textbf{0.903 } & 0.846  & \textbf{0.902 } & \textbf{0.915 } & \textbf{0.947 } & \textbf{0.025 } \\
    \hline
    \multicolumn{1}{c|}{\multirow{7}{*}{CVC-ClinicDB}}
          & w/o LFM (2) & 0.887  & 0.827  & 0.887  & 0.915  & 0.951  & 0.013  \\
          & w/o PAM & 0.877  & 0.812  & 0.890  & 0.912  & 0.952  & 0.013  \\
          & PAM (PA-RA) & 0.909  & 0.862  & 0.910  & 0.931  & 0.963  & 0.008  \\
          & PAM (PA-BA) & 0.904  & 0.846  & 0.904  & 0.923  & 0.962  & 0.012  \\
          & w/o BWM & 0.898  & 0.850  & 0.905  & 0.925  & 0.955  & 0.008  \\
          & \textbf{Ours} & \textbf{0.918 } & \textbf{0.869 } & \textbf{0.927 } & \textbf{0.935 } & \textbf{0.971 } & \textbf{0.008 } \\
    \hline
    \end{tabular*}%
  \label{tab:5}%
\end{table*}%

\begin{table*}[htbp]
  \centering
  \caption{Ablation study on the CVC-300, CVC-ColonDB and ETIS datasets.}
    \begin{tabular*}{\linewidth}{m{1cm}<{\centering}|m{2.5cm}<{\centering}m{1.9cm}<{\centering}m{1.9cm}<{\centering}m{1.9cm}<{\centering}m{1.9cm}<{\centering}m{1.9cm}<{\centering}m{1.5cm}<{\centering}}
    
    \hline
    \multicolumn{1}{c|}{\textbf{Dataset}}
          & {\textbf{Settings}} & {\textbf{meanDice}} & {\textbf{meanIoU}} & {\textbf{$\bm{F_\beta^\omega}$}} & {\textbf{$\bm{S_m}$}} & {\textbf{$\bm{E_\phi^{\max}}$}} & {\textbf{MAE}} \\
    \hline
        
    \multicolumn{1}{c|}{\multirow{7}{*}{CVC-300}} 
          & w/o LFM (2) & 0.872  & 0.804  & 0.858  & 0.915  & 0.957  & 0.011  \\
          & w/o PAM & 0.861  & 0.793  & 0.852  & 0.910  & 0.939  & 0.008  \\
          & PAM (PA-RA) & 0.848  & 0.769  & 0.821  & 0.898  & 0.943  & 0.012  \\
          & PAM (PA-BA) & 0.879  & 0.810  & 0.880  & 0.924  & 0.958  & 0.007  \\
          & w/o BWM & 0.851  & 0.784  & 0.840  & 0.906  & 0.938  & 0.009  \\
          & \textbf{Ours} & \textbf{0.907 } & \textbf{0.842 } & 0.893  & \textbf{0.935 } & \textbf{0.980 } & \textbf{0.005 } \\
    \hline
    \multicolumn{1}{c|}{\multirow{7}{*}{CVC-ColonDB}} 
          & w/o LFM (2) & 0.743  & 0.663  & 0.734  & 0.826  & 0.872  & 0.036  \\
          & w/o PAM & 0.698  & 0.617  & 0.699  & 0.799  & 0.829  & 0.043  \\
          & PAM (PA-RA) & 0.738  & 0.663  & 0.728  & 0.824  & 0.867  & 0.039  \\
          & PAM (PA-BA) & 0.710  & 0.628  & 0.710  & 0.806  & 0.843  & 0.041  \\
          & w/o BWM & 0.745  & 0.674  & 0.737  & 0.829  & 0.876  & 0.037  \\
          & \textbf{Ours} & \textbf{0.762} & \textbf{0.690} & \textbf{0.754} & \textbf{0.838} & \textbf{0.892} & \textbf{0.036 } \\
    \hline
    \multicolumn{1}{c|}{\multirow{7}{*}{ETIS}}
          & w/o LFM (2) & 0.590  & 0.512  & 0.537  & 0.753  & 0.743  & 0.063  \\
          & w/o PAM & 0.614  & 0.539  & 0.596  & 0.768  & 0.834  & 0.019  \\
          & PAM (PA-RA) & 0.584  & 0.508  & 0.540  & 0.744  & 0.754  & 0.057  \\
          & PAM (PA-BA) & 0.704  & 0.623  & 0.673  & 0.819  & 0.864  & 0.016  \\
          & w/o BWM & 0.597  & 0.522  & 0.560  & 0.758  & 0.784  & 0.029  \\
          & \textbf{Ours} & \textbf{0.764 } & \textbf{0.686 } & \textbf{0.739 } & \textbf{0.855 } & \textbf{0.900 } & \textbf{0.012 } \\
    \hline
    \end{tabular*}%
  \label{tab:6}%
\end{table*}%

\begin{figure*}  
\centering 
\includegraphics[width = \textwidth]{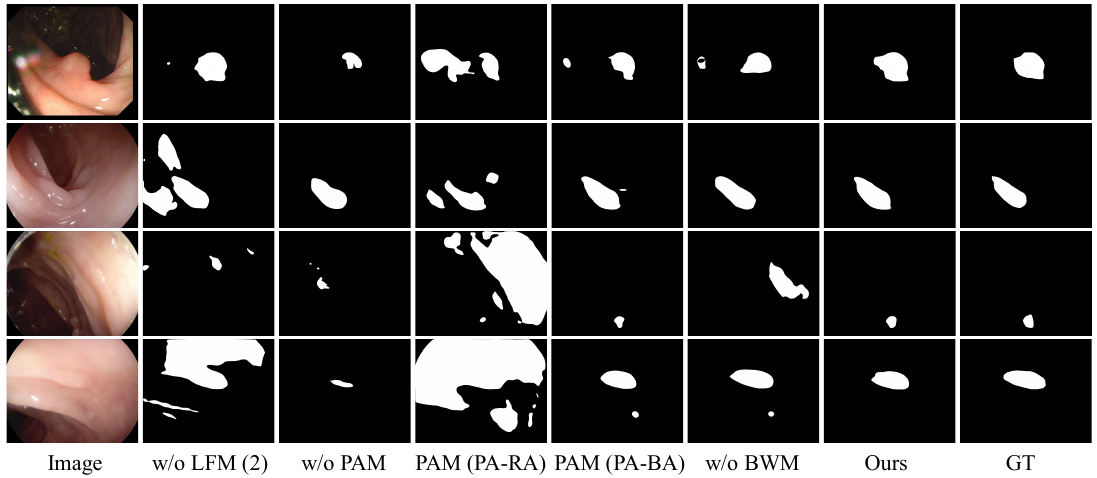} 
\caption{Visual comparisons about ablation study.}
\label{fig_10}
\end{figure*}

\textbf{Effectiveness of Parallel Attention Module.} 
To evaluate the effectiveness of Parallel Attention Module, we investigate the following settings:
(1) w/o PAM: we remove PAM from the network, so that each side feature map is generated only by fusing deep features with previous layer features through the balancing weight module. 
(2) PAM (PA-RA): we only apply reverse attention.
(3) PAM (PA-BA): we only apply boundary attention.

The results in Table \ref{tab:5} and Table \ref{tab:6} proved the effectiveness of PAM module. Meanwhile, we can see that with only PA-RA or PA-BA, the model fail to reach the best performance, demonstrating that these two modules provide attention to the boundary from different perspectives and their combination is the most effective setting. Figure \ref{fig_10} also indicates that adequate attention to boundaries can improve the segmentation performance. 





\textbf{Effectiveness of Balancing Weight Module.} To validate the effectiveness of the balancing weight module, we conduct ablation experiments by replace BWM with direct adding operation(denoted as w/o BWM). The results in Table \ref{tab:5}, Table \ref{tab:6} and the visualization result in Figure \ref{fig_10} demonstrates that the approach of merging features by balancing the weights of multiple sources enhances the segmentation performance. With BWM, the network forms a layer-by-layer architecture to progressively refines the details and incorporates semantic information from deep levels, and finally obtain the output segmentation results.

\section{Conclusion}

In this paper, we propose a novel network architecture MISNet for polyp segmentation
Notably, a Selective Shared Fusion Module is proposed to promote information sharing and active selection across various feature levels to enhance the model's ability to capture multi-scale contextual information, improving the accuracy of the initial guidance map used in the decoding stage.
A Parallel Attention Module is proposed to emphasize the model's attention to boundary information. Meanwhile, a Balancing Weight Module embedded in the bottom-up flow allows for adaptive integration of features from different layers.
Experimental results on five polyp segmentation datasets show that the proposed method outperforms state-of-the-arts under different metrics.

In future work, we will focus on semi-supervised and self-supervised learning approaches to address issues such as insufficient data volume and class imbalance. We plan to integrate multi-modal information, such as CT and MRI, to provide more comprehensive and accurate medical image information. Additionally, we will explore how to adopt lightweight model structures without compromising segmentation accuracy.


\bibliographystyle{ieeetr}
\normalem
\bibliography{ref} 

\end{document}